\pdfoutput=1

\documentclass[11pt]{article}

\usepackage{EMNLP2022}

\usepackage{times}
\usepackage{latexsym}

\usepackage[T1]{fontenc}

\usepackage[utf8]{inputenc}

\usepackage{microtype}

\usepackage{inconsolata}

\usepackage{amsmath}
\usepackage{cleveref}
\usepackage{tabularx}

\usepackage{colortbl}
\usepackage{graphicx}
\usepackage{amssymb}
\usepackage{booktabs}
\usepackage{multirow}
\usepackage{multicol}

\usepackage{algpseudocode}
\usepackage{algorithm}

\usepackage{caption,subcaption}
\usepackage{pgfplots}
\pgfplotsset{compat=1.17}

\newcommand{\dataset}{{\sc CovidET}}

\usepackage[textsize=scriptsize]{todonotes}

\newcommand{\secref}[1]{\S\ref{#1}} 
\usepackage{makecell} 
\usepackage{adjustbox,lipsum}
\usepackage{listings}
\title{Why Do You Feel This Way? \\ Summarizing Triggers of Emotions in Social Media Posts}

 \author{
    \textbf{Hongli Zhan$^*$}$^1$\quad\textbf{Tiberiu Sosea$^*$}$^2$\quad\textbf{Cornelia Caragea}$^2$\quad\textbf{Junyi Jessy Li}$^1$\\
    $^1$Department of Linguistics, The University of Texas at Austin\\$^2$Department of Computer Science, University of Illinois Chicago\\
    \texttt{\{honglizhan,jessy\}@utexas.edu}\quad\texttt{\{tsosea2,cornelia\}@uic.edu}
}

\begin{document}
\maketitle

\begingroup\begin{NoHyper}\def\thefootnote{*}\footnotetext{Hongli Zhan and Tiberiu Sosea contributed equally.}\end{NoHyper}\endgroup

\begin{abstract}
    Crises such as the COVID-19 pandemic continuously threaten our world and emotionally affect billions of people worldwide in distinct ways. Understanding the triggers leading to people's emotions is of crucial importance. Social media posts can be a good source of such analysis, yet these texts tend to be charged with multiple emotions, with triggers scattering across multiple sentences. This paper takes a novel angle, namely, \textit{emotion detection and trigger summarization}, aiming to both detect perceived emotions in text, and summarize events and their appraisals that trigger each emotion. To support this goal, we introduce \dataset{} (\emph{\textbf{E}motions and their \textbf{T}riggers during \textbf{Covid}-19}), a dataset of \textasciitilde$1,900$ English Reddit posts related to COVID-19, which contains manual annotations of perceived emotions and abstractive summaries of their triggers described in the post. We develop strong baselines to jointly detect emotions and summarize emotion triggers. Our analyses show that \dataset{} presents new challenges in emotion-specific summarization, as well as multi-emotion detection in long social media posts.
\end{abstract}

\section{Introduction}
    
\begin{figure}
    \centering
    \includegraphics[width=0.95\columnwidth]{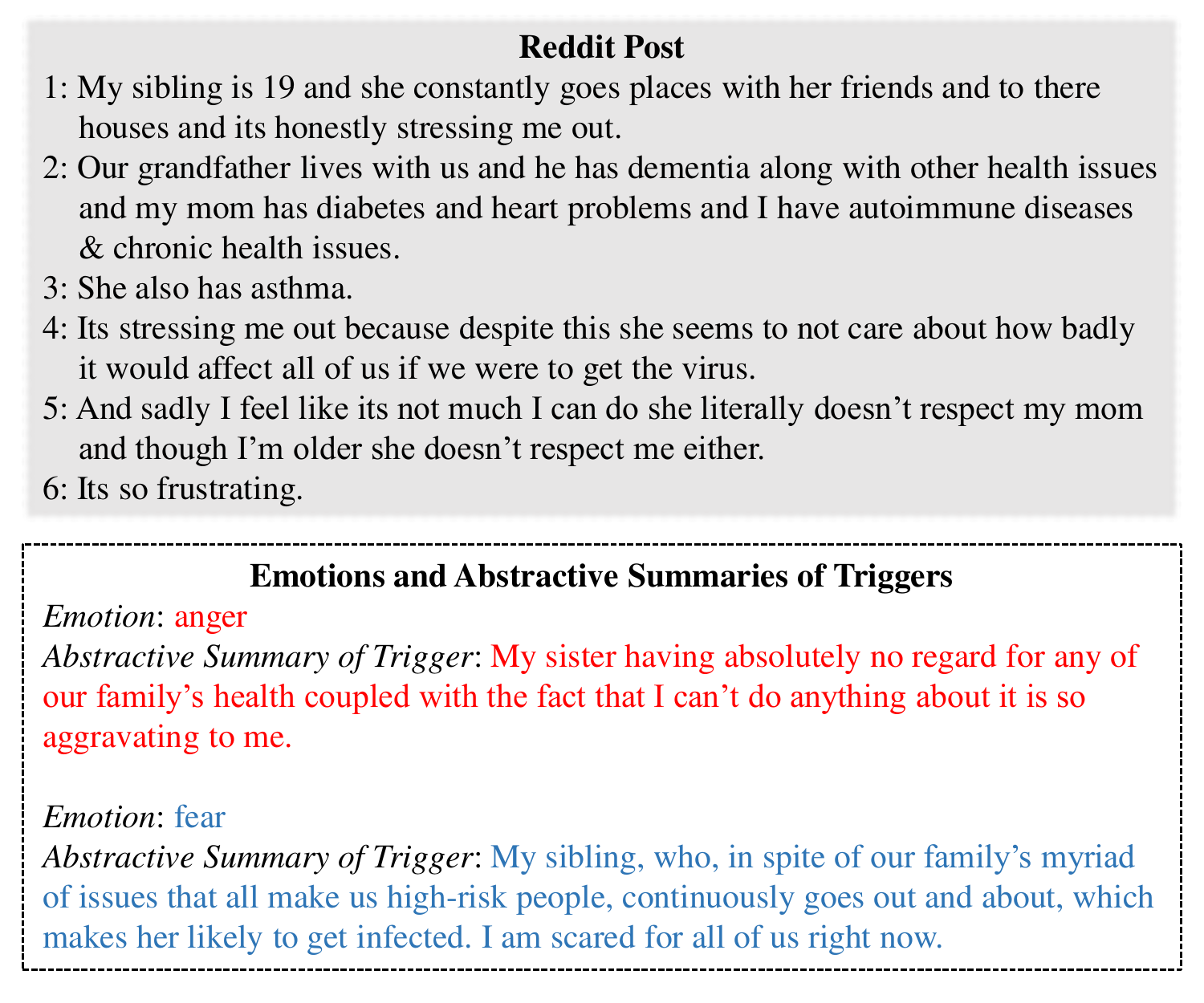}
    \vspace{-0.5em}
    \caption{An example from \dataset{}, with perceived emotion(s) identified and their trigger(s) summarized.}
    \label{fig:dataset-example}
\end{figure}

Large-scale crises such as the COVID-19 pandemic continuously cause emotional turmoil worldwide. People are emotionally affected in different ways, e.g., online education has led to mental health issues among students~\cite{akpinar2021effect} as well as parents~\cite{info:doi/10.2196/24496}; lock-down policies are protective for the vulnerable~\cite{flaxman_estimating_2020, hsiang_effect_2020} while economically disastrous for many~\cite{Odii_Ngwu_Aniakor_Owelle_Aniagboso_Uzuanwu_2021}. Emotion analysis --- both \emph{detecting} emotion and understanding what \emph{triggers} the emotion --- brings invaluable insights both practically (e.g., for first responders, counselors, etc) and in scientific research~\cite{arora_role_2021, UBAN2021480}.

While emotion detection (typically formulated as a classification task among standard emotion taxonomies) is a well-established task (e.g., \citet{mihalcea-strapparava-2012-lyrics, Wang2012HarnessingT, abdul-mageed-ungar-2017-emonet, khanpour-caragea-2018-fine, demszky-etal-2020-goemotions} and in crises contexts~\cite{desai-etal-2020-detecting, sosea2021emotion}), fewer have studied \textbf{\textit{what leads to these emotions}}
in the scope of the text concerned in a data-driven manner. \citet{xia-2019-ECPE} adopt an extraction setup 
to identify emotion ``causes''
that is limited to the clause level, where only one (explicitly expressed) emotion and one cause are associated. This does not generalize to long, spontaneous social media posts that are emotionally charged. Illustrated in Figure~\ref{fig:dataset-example}, distinct emotions are triggered by different events across multiple sentences. 

Additionally, how these events are subjectively evaluated, interpreted or \emph{appraised}, e.g., ``I can't do anything about it'' in the first example of Figure~\ref{fig:dataset-example}, also contribute to the emotion~\cite{smith1985patterns, ellsworth2003appraisal}. The fact that different individuals may have distinct appraisals towards the same event \cite{moors-appraisal-2013} further highlights the challenging nature of understanding what triggers an emotion.

In this work we take a novel view, and formulate emotion-trigger detection as an \emph{abstractive summarization} task that synthesizes a natural language description of the events and their appraisals that trigger a particular emotion. We frame our work as \emph{emotion detection and trigger summarization} (Figure~\ref{fig:dataset-example}), which entails both detecting perceived emotions in text, and summarizing triggers for each emotion.

We present \dataset{} (\emph{\textbf{E}motions and their \textbf{T}riggers during \textbf{Covid}-19}), a new dataset sourced from $1,883$ English Reddit posts about the COVID-19 pandemic. Each post is annotated with 7 fine-grained emotion labels; for each emotion, annotators provided a concise, abstractive summary describing the triggers of the emotion. 
The triggers are further \emph{validated} in a separate stage. \dataset{} spans from June 2021 to January 2022, capturing various significant events as well as how they were emotionally appraised during the pandemic. Compared to prior emotion studies that consider only sentence-level texts \cite{sosea-caragea-2020-canceremo,demszky-etal-2020-goemotions} or (short) tweets \cite{sosea2021emotion,abdul-mageed-ungar-2017-emonet}, \dataset{} is challenging as it contains significantly longer texts. We showcase examples of \dataset{} in Appendix \secref{appendix:dataset-examples}.

Analyses of \dataset{} reveal that negative emotions such as \textit{fear} and \textit{anger} are prevalent. These emotions co-occur most frequently with \textit{anticipation}, which consistently rise after the Omicron subvariant became more dominant with \textit{fear} dropping. Topic modeling over the trigger summaries points to irritations toward those who don't mask or get vaccinated, and positivity towards the vaccines.

Using \dataset{}, we benchmark models for emotion detection and emotion-trigger summarization. We employ both separate emotion detection and trigger summarization models, as well as joint models that we designed to simultaneously detect emotions and generate trigger summaries. Our experiments showcase the distinct nature of our task, emphasizing that \dataset{} is vital to training reliable emotion detection and trigger summarization approaches in a Covid-19 context. \dataset{} bears various unique characteristics, ranging from its long sequences and invaluable context to the nature of the task itself. Therefore, general emotion detection or summarization models unsurprisingly lag behind in performance compared to our methods. Moreover, human evaluation of the generated trigger summaries tailored for emotion-trigger summarization indicates that our models are effective in capturing the underlying triggers of the post.

We release \dataset{} and our code at \url{https://github.com/honglizhan/CovidET}.

\section{Related Work}

\paragraph{Summarization.}
Recent pre-trained models led to substantial progress in single document summarization. In the case of abstractive summarization, encoder-decoder transformer models are used to synthesize a concise description of the most salient concepts in the input~\cite{lewis-etal-2020-bart,pmlr-v119-zhang20ae}. Significant efforts in summarization focus on news because of the availability of large datasets such as CNN/DailyMail~\cite{nips15_hermann} and XSum~\cite{narayan-etal-2018-dont}; in the domain of social media, TL;DR sentences has been mined in Reddit to serve as summaries and train models~\cite{volske-etal-2017-tl,Kim:2019:NAACL-HLT}. However, generic summaries tend not to be informative if users are concerned with specific emotions expressed.

In this sense our setup fits into settings where only a certain part of the content is of interest to the user. We could view our task as answering a query, ``\emph{Why does the writer feel [emotion]?}''. However, such queries are more general than query-based summarization~\cite{daume-iii-marcu-2006-bayesian, OTTERBACHER200942, schilder-kondadadi-2008-fastsum, nema-etal-2017-diversity,DBLP:journals/corr/abs-1801-07704, Laskar2020QueryFA, su-etal-2021-improve, zhong-etal-2021-qmsum}, where queries tend to be more document-specific. Perhaps a closer task is opinion summarization, or aspect-based summarization more generally. In opinion summarization, models need to summarize affect/opinions about a certain aspect of a service or product~\cite{popescu-etzioni-2005-extracting, angelidis-lapata-2018-summarizing, huy-tien-etal-2019-opinions, suhara-etal-2020-opiniondigest, angelidis-etal-2021-extractive, amplayo-lapata-2021-informative}; on the contrary, our setup entails identifying the emotions and summarizing the events and how they were made sense of with respect to each emotion. In aspect-based summarization, existing work has explored summarizing with respect to pre-designated aspects of certain news~\cite{frermann-klementiev-2019-inducing, ahuja-etal-2022-aspectnews}, and entities mentioned in  text~\cite{maddela-etal-2022-entsum}.

\paragraph{Emotion Cause Extraction.}
Emotion Cause Extraction (ECE) is a task that aims to extract the events triggering a particular emotion \cite{khunteta-2021-review}. ECE was first introduced by \citet{lee-etal-2010-emotion}, where they defined the task as extracting word-level causes to the given emotion in text. \citet{chen-etal-2010-emotion} and \citet{gui-etal-2016-event} expanded the task to clause-level cause detection; \citet{xia-2019-ECPE} aimed to removed the constraint that emotions must be human-annotated before conducting automatic cause extraction, and thus proposed  Emotion-Cause Pair Extraction (ECPE) aiming to extract potential pairs of emotions and causes in a document. Most of the datasets are in Chinese, in either micro-blog or news domains \cite{GAO20154517, gui-etal-2016-event, Gao2017OverviewON}.

In contrast, we study a more generalized notion of \emph{triggers} of an emotion where readers are asked to actively appraise and interpret the emotions together with their stimuli in the document, rather than solely identifying the events behind each emotion. We use abstractive summarization to handle triggers, which can better synthesize inter-connected complex events and abstract concepts, as well as making the output contextually independent.

\section{Dataset Construction}\label{sec:Dataset-Construction}
    We present \dataset{}, a novel dataset from English Reddit posts that is manually annotated with emotions and summaries of their triggers. This section discusses the data creation process; in \S\ref{sec:Inter-Annotator-Agreement}, we discuss inter-annotator agreement and our human verification process.

    \subsection{Selecting \& Curating Reddit Posts}\label{subsec:data-selecting}
    We gather posts from  \texttt{r/COVID19\_support}\footnote{\url{https://www.reddit.com/r/COVID19_support/}}. We select it as the source of our data because of its rich personal narration: rather than COVID-19 news snippets, this subreddit is targeted for people seeking any community support during the pandemic. We randomly sample posts before (from Jun 23, 2021 to Oct 1, 2021) and after (from Dec 1, 2021 to Jan 25, 2022) Omicron, a COVID-19 variant that emerged during December 2021.
    
\begin{figure}[!t]
    \centering
    \includegraphics[width=0.8\linewidth]{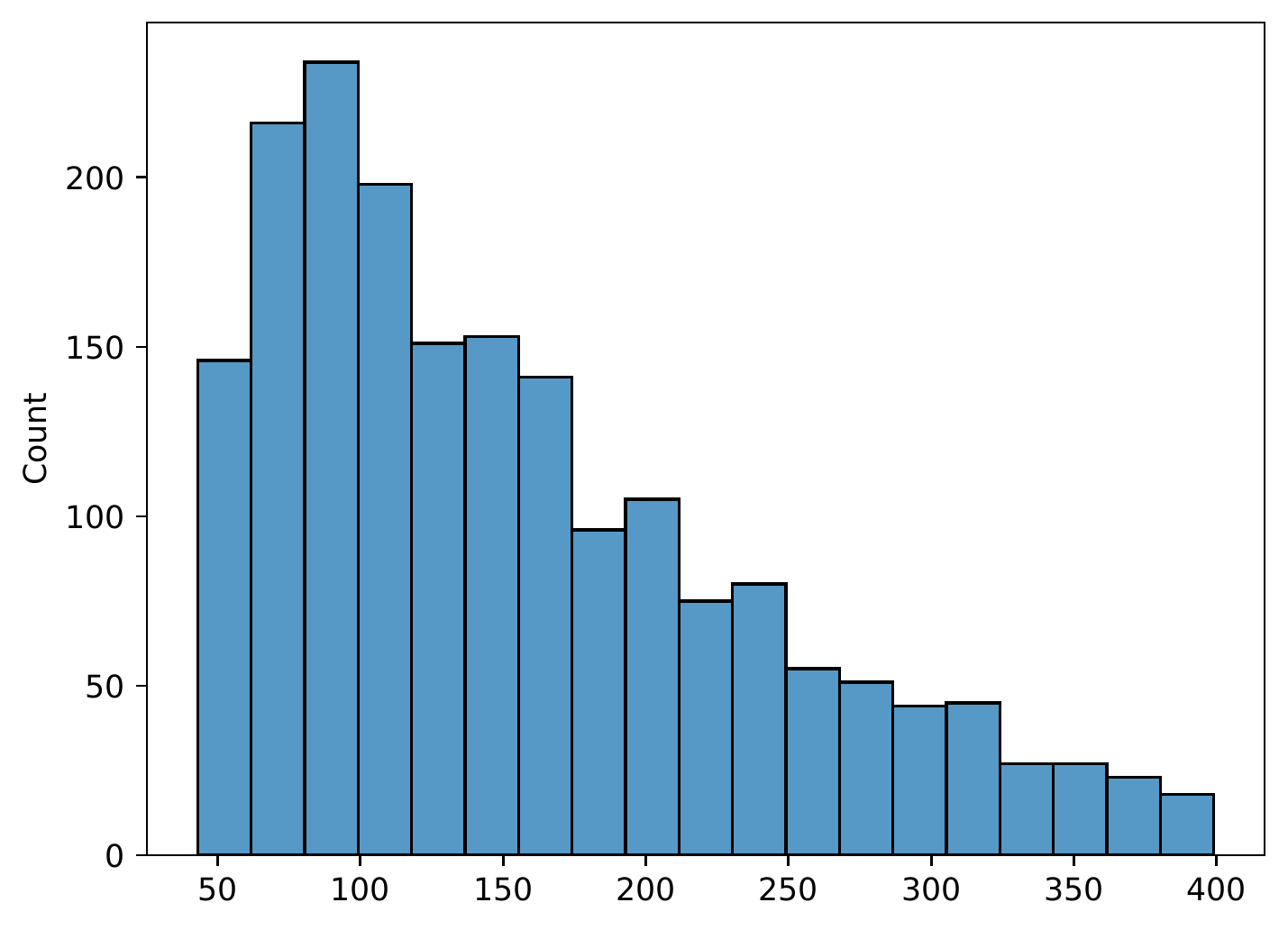}
    \caption{Distribution of the length of posts in \dataset{}.}
    \label{fig:post_length}
\end{figure}

    We restrict posts to be between $50$-$400$ tokens long (punctuation excluded); this allows us to have posts that are long enough, but still manageable for crowdsourcing tasks. Close scrutiny shows that the posts in \dataset{} center around 100 tokens in length; the distribution of the length of the posts is given in Figure \ref{fig:post_length}. The average length of posts in \dataset{} is $156.4$ tokens (std.dev = $83.3$). We mask web links with an \texttt{[url]} token and do not provide the metadata to our annotators. Note that 6 posts have length under 50 tokens: this is because we performed \texttt{[url]} masking \textit{after} length filtering when collecting the source data. Details of the full preprocessing procedure are provided in Appendix \secref{appendix:data-curation}.

    \subsection{Annotation Task}\label{subsec:data-annotation}
        \paragraph{Instructions.}
            Annotators are first asked to annotate Plutchik basic emotions~\cite{plutchik2001the} they perceive: \textit{anger, anticipation, joy, trust, fear, sadness, and disgust}.\footnote{After annotation, we found very little \emph{surprise} in the data (59 in total), thus we leave out \emph{surprise} for this work.} Multiple selection is allowed, and we also provide a \textit{none of the above} option in case no emotion is perceived.

            Once the annotators select an emotion, they are asked to summarize the trigger(s) to their perceived emotions, specifically an abstractive summary in their own words, in the author's voice. The summaries should contain trigger(s) to the emotion rather than just reflecting the emotion itself. We provide the detailed instructions to our annotation task in Appendix \secref{appendix-subsec:annotation-instructions}.

        \paragraph{Annotators}
            We recruit two different groups of annotators. The first group consists of trained turkers from Amazon Mechanical Turk. The workers' locale is the US, and they have completed $500$+ HITs with an acceptance rate $\geq$ $95$\%. This group contributes to \dataset{}'s training and validation sets. The second group consists of 2 linguistic undergraduate students, who contributes to the test set. To ensure the quality of \dataset{}, both groups of annotators are trained and qualified in a pre-qualification process. We also ask them to revise their work when needed during annotation.
            
    \paragraph{Pre-Annotation Training}
        We trained the annotators before they annotate \dataset{}. We set up a qualification task on the Amazon Mechanical Turk. The qualification task involves 3 posts, and annotators are required to complete the qualification task. Through manually examining the annotators' work on the qualification task and comparing the annotations to the \textit{gold} annotations we develop, we filter high-quality annotators and give them the access to our annotation task. We also provide feedback to their annotations. The turkers are paid at least \$10 per hour. To ensure this goal is reached, we keep track of their working time on the backstage and give out bonuses accordingly when needed.
    
    \paragraph{Annotation Revisions}
        During the process of the annotation on \dataset{}, we regularly review the annotations and give feedback accordingly. When needed, we send the annotations back to the annotator along with the original post, and ask them to revise their work based on our suggestions. Note that the annotator is responsible for the revision of \textit{their own} work only.

\subsection{Benchmark Dataset}\label{subsec:benchmark-dataset}
We annotated $1,485$ posts on the Amazon Mechanical Turk, each annotated by two independent workers. Since the neutral class is very infrequent, we remove it from our experiments. To facilitate our experiments, we split the examples into $1,200$ examples for training and $285$ examples for validation. Our test set---which is annotated by linguistic undergraduates---contains $398$ examples.

If at least one annotator labels a post with an emotion $e$, then we include emotion $e$ as an emotion label. In cases where both annotators assign an emotion $e$ to a post, we consider the trigger summaries as two separate training examples for the trigger summarization task. In cases where a post has two different trigger summaries in the validation or the test set, we consider them as multiple references when computing our metrics.

\section{Agreement and Validation}\label{sec:Inter-Annotator-Agreement}
    To account for the quality of \dataset{}, we measure the inter-annotator agreement in emotions (\secref{subsec:agreement-emotions}) and triggers (\secref{subsec:agreement-triggers}). The annotations are further validated through human inspection in \secref{subsec:human-validation}. Results reveal that annotators tend to agree with each other in emotions whilst using varied vocabularies when summarizing the triggers.
    
            \begin{figure}[t]
                \centering
                \includegraphics[width=0.9\columnwidth]{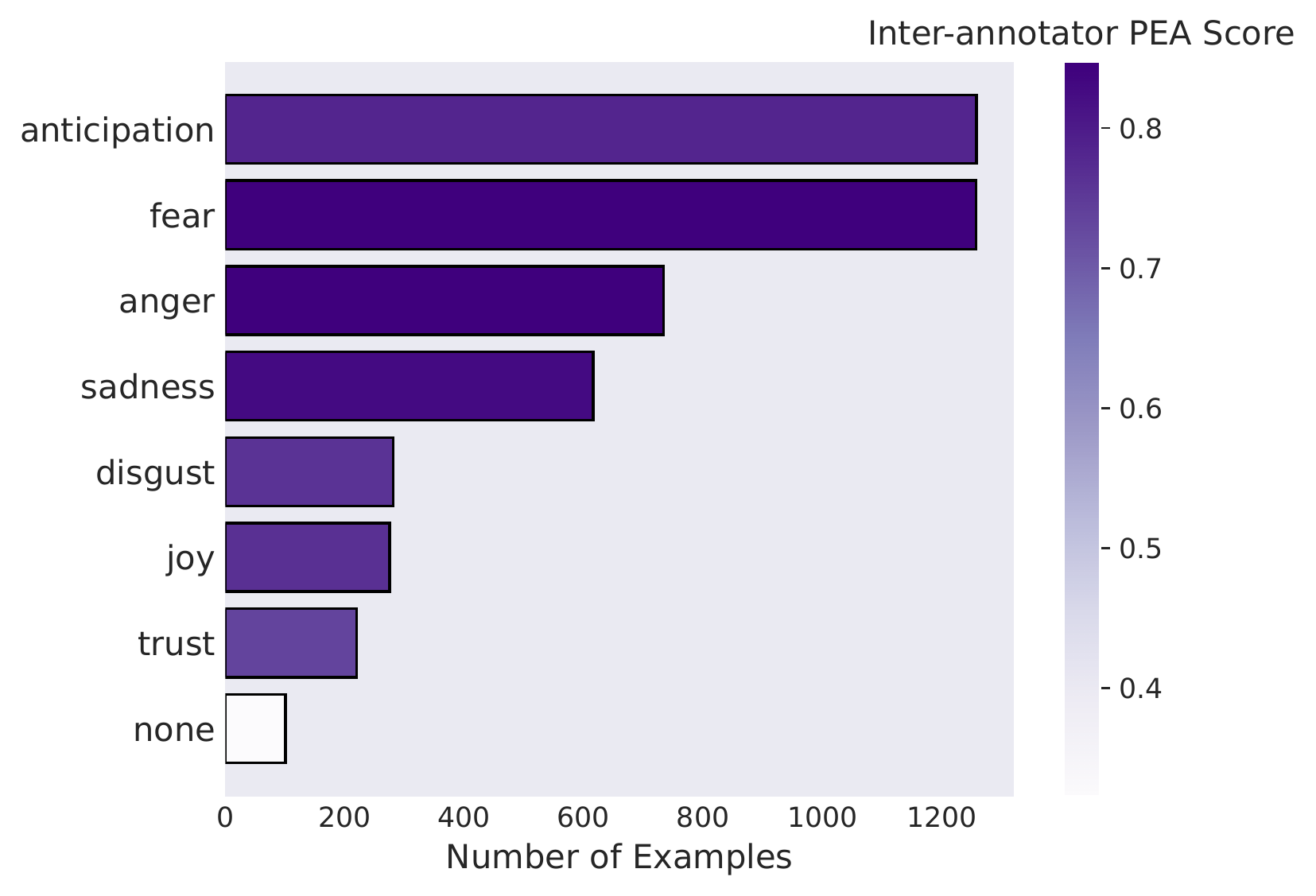}
                \caption{Emotion distribution of \dataset{}, ranked by the number of examples. Colors indicate the inter-annotator agreement measured by the PEA score.}
                \label{fig:emotion-distribution}
            \end{figure}
            
    \subsection{Agreement in Emotions}\label{subsec:agreement-emotions}
        \paragraph{Percentage Overlap.} For each example in \dataset{}, we measure the number of emotions in which both annotators agree upon. Results show that in 81.4\% of the examples in \dataset{}, both annotators agree on at least 1 emotion label; in 26.6\% of the examples, both annotators agree on at least 2 emotion labels.
        
        \paragraph{PEA Score.} To account for distances between emotions (e.g., disgust is further away from joy than from anger), we report the Plutchik Emotion Agreement (PEA) metric \cite{desai-etal-2020-detecting} for the inter-annotator agreement of emotions annotated in \dataset{}. We first report the average PEA score among annotators weighted by their numbers of annotations, which is 0.8 for the training and validation sets combined, and 0.821 for the test set (0.804 for all three combined). These numbers indicate high agreement~\cite{desai-etal-2020-detecting}.
            
        Figure \ref{fig:emotion-distribution} shows per-emotion PEA scores, along with the frequency of each emotion. All emotions have high agreement; the highest are among \textit{fear} and \textit{anger}, with the average PEA scores at around 0.85; the lowest is \textit{trust}, with the average PEA score at around 0.74.

        Finally, to calculate agreement between students and crowd workers, we randomly sample 208 examples from the training set and ask the linguistic undergraduate students to annotate them from scratch. We assign one student per example for validation. The average PEA score between crowd workers and linguistics students is 0.832, suggesting high agreement.

    \subsection{Similarity in Triggers}\label{subsec:agreement-triggers}
        We further examine the similarity in the annotated summaries of triggers when two annotators both select the same emotion for one example, using ROUGE~\cite{lin-2004-rouge} for lexical overlap and BERTScore~\cite{DBLP:journals/corr/abs-1904-09675} for semantic similarity. The average BERTScore (F1) between the two annotators is 0.883, indicating highly similar summaries. Yet the lexical overlap is low:
        the average ROUGE F scores between two annotators are: ROUGE-1: 0.255, ROUGE-2: 0.055, ROUGE-L: 0.190.

        For those posts doubly annotated by linguistics students and crowd workers, the ROUGE values are similar for students vs.\ workers: BERTScore: 0.876; ROUGE-1: 0.246, ROUGE-2: 0.063, ROUGE-L: 0.188. \footnote{Multi-reference ROUGE and BERTScore are applied in cases where all 3 annotators agree in the same emotion.}
    
        \begin{table}
            \centering
              \setlength{\tabcolsep}{2pt}
              \small
                \begin{tabular}{l|ccccccccc}
                    \toprule
          & AGR   & DSG   & FER   & JOY   & SDN   & TRS   & ANC   & Avg \\
                    \midrule
    Emotion & 0.96  & 0.92  & 0.96  & 1     & 0.96  & 0.88  & 0.92  & 0.94 \\
    Trigger & 0.92  & 0.92  & 0.96  & 1     & 0.96  & 0.84  & 0.88  & 0.93 \\
                    \bottomrule
                \end{tabular}
            \caption{Human validation results on the annotated emotions and abstractive summaries of triggers.}
            \label{tab:dataset-validation-abst}
        \end{table}
    
    \subsection{Human Validation}\label{subsec:human-validation}
        In addition to the automatic evaluation metrics above, we also validate the emotion-trigger annotations in \dataset{} through human inspections. We set up a human validation task on the Amazon Mechanical Turk, and recruit a new group of qualified workers. We randomly sample 300 examples from our training set for validation. The emotion annotations, as well as the abstractive summaries of triggers, are validated.
        
        We describe the validation framework as follows. The validators are given an annotated trigger summary. We first validate whether the summary actually indicates the annotated emotion by asking a yes/no question. Next, if the validator confirms the presence of emotion in the summary, we then ask whether the summary indeed expresses the \textit{trigger} and not the \textit{emotion} by raising another yes/no question. We present the validation results based on the abstractive summaries in Table \ref{tab:dataset-validation-abst}. The numbers indicate the proportion of examples on which validators confirm upon.
        
        \begin{table}[t]
  \setlength{\tabcolsep}{2pt}
  \centering
  \tiny
    \begin{tabular}{c|c|c|c|c|c|c}
    \toprule
    AGR   & DSG   & FER   & JOY   & SDN   & TRS   & ANC \\
    \midrule
    covid & disgusted & covid & happy & sad   & trust & covid \\
    annoyed & covid & afraid & covid & covid & covid & expect \\
    people & people & getting & vaccinated & feel  & vaccine & looking \\
    angry & feel  & scared & pandemic & pandemic & information & know \\
    don   & don   & going & vaccine & life  & people & interested \\
    vaccinated & pandemic & vaccine & getting & don   & believe & people \\
    pandemic & getting & worried & people & like  & vaccines & symptoms \\
    just  & like  & risk  & feel  & just  & help  & test \\
    want  & vaccine & concerned & better & people & pandemic & getting \\
    life  & just  & health & good  & friends & vaccinated & want \\
    going & mask  & vaccinated & able  & time  & credible & going \\
    vaccine & going & symptoms & really & lost  & protect & vaccinated \\
    getting & vaccinated & fear  & know  & going & know  & vaccine \\
    family & want  & don   & news  & really & end   & positive \\
    really & family & effects & vaccination & want  & feel  & guidance \\
    \bottomrule
    \end{tabular}%
  \caption{Results of topic modelling through LDA \cite{10.5555/944919.944937}. The words are associated with the most prominent topic among the abstractive summaries of triggers of each emotion category in \dataset{}.}
  \label{tab:lda}%
\end{table}
            \begin{figure}
                \centering
                \includegraphics[width=\columnwidth]{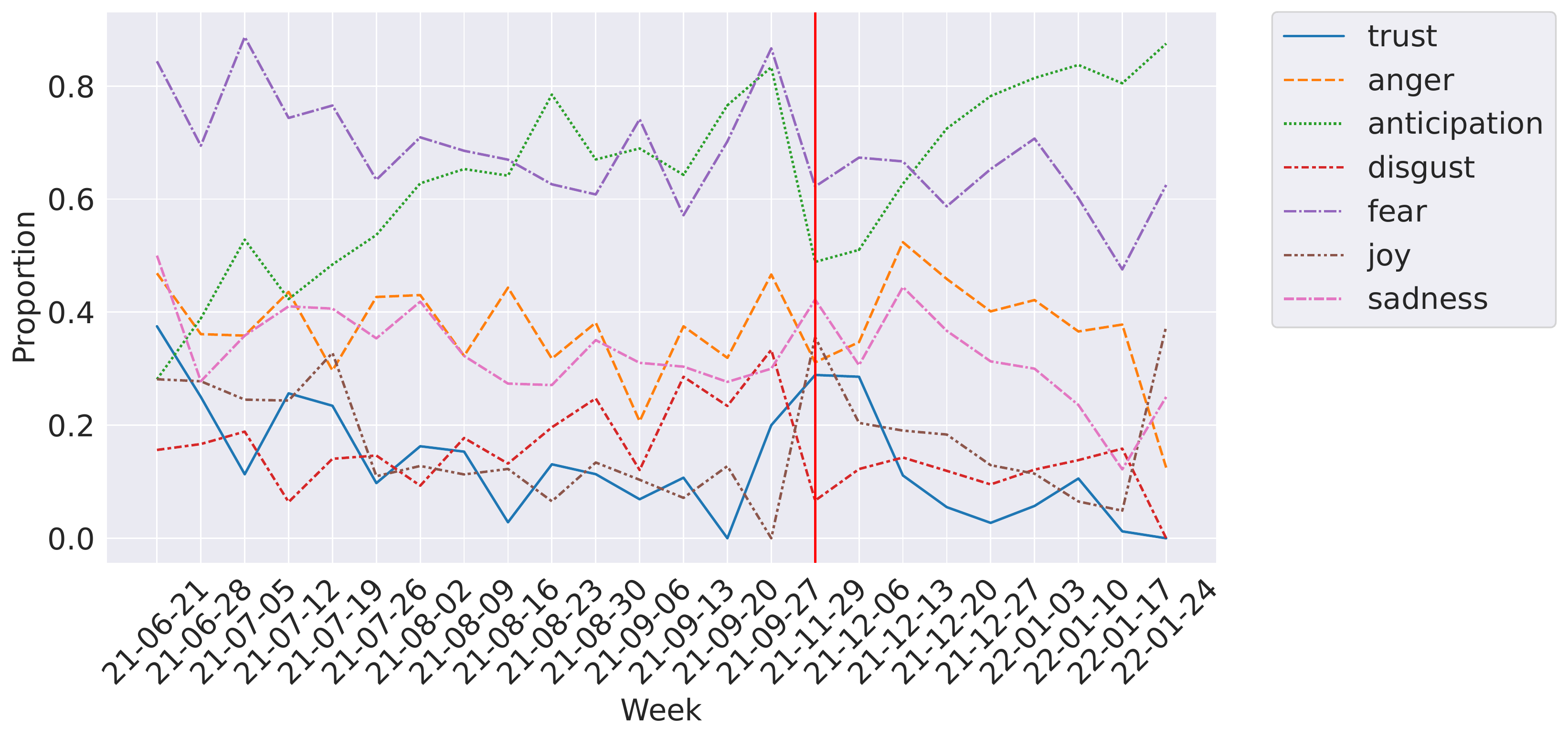}
                \caption{Emotion distribution in \dataset{} over time (by week).}
                \label{fig:trend}
            \end{figure}
        
        Overall, the human validation results indicate fairly high correctness in our annotations. It should be noted that annotators commonly adopt some sentence patterns that can be easily identified as emotion triggers. For example, in expressing the abstractive trigger for \textit{anger}, an annotation in \dataset{} is \textit{I am angry that they would put me at risk of catching COVID and not tell me}, a sentence which is highly linguistically explicit of the emotion.

\section{Data Analysis}\label{sec:Analysis}

        \paragraph{Emotion Distribution.}
        On average, there are $2.46$ emotions (``none'' excluded) per example in \dataset{}. Figure \ref{fig:emotion-distribution} shows the general emotion distribution of \dataset{}. \textit{Fear} is the most common emotion in \dataset{}, closely followed by \textit{anticipation}. There is clearly a gap among the emotions, with positively valenced emotions such as \textit{trust} and \textit{joy} rarely present in \dataset{}. This is predicted given the catastrophic nature of our domain.

\begin{figure}[b]
    \centering
    \includegraphics[width=0.9\columnwidth]{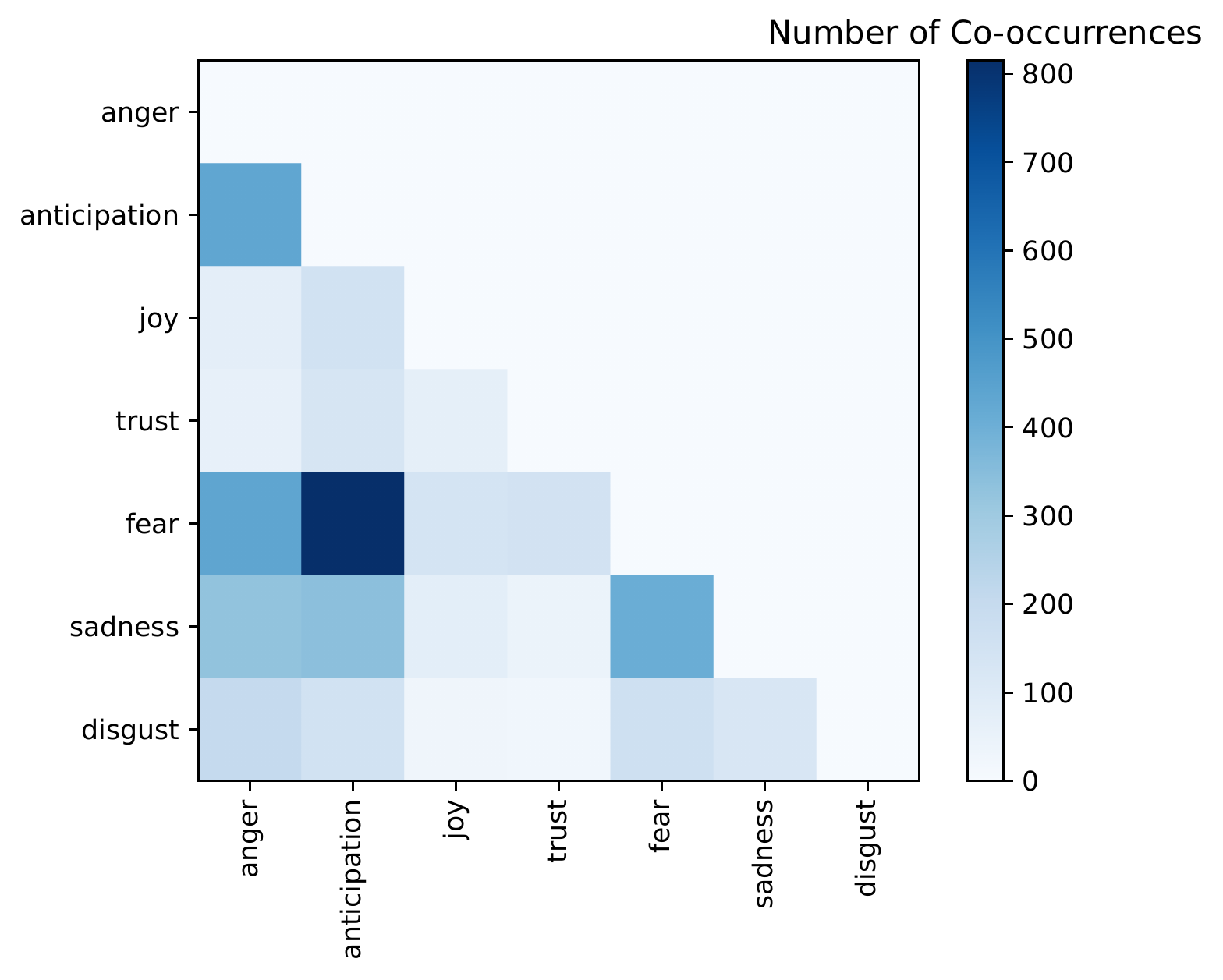}
    \caption{Emotion co-occurrences in \dataset{}.}
    \label{fig:heatmap}
\end{figure}

        We present the emotion co-occurrence heatmap in Figure \ref{fig:heatmap}. \textit{Anticipation} co-occurs with \textit{fear} and \textit{anger} most frequently in \dataset{}. Close scrutiny of the data reveals that the poster is either predicting negative events during COVID-19, or expecting advice on what do do under austere situations.

        \paragraph{Emotion Trend.}
            We present a temporal analysis of the emotion distribution by week in Figure \ref{fig:trend}, using a red vertical line to separate pre- and post-Omicron. Interestingly, we notice that the amount of \textit{anticipation} consistently rises after the outbreak of the Omicron COVID-19 variant, whereas the expression of negative emotions including \textit{anger} and \textit{fear} becomes less prevalent, possibly due to the nature of the Omicron variant, which was less harmful compared to previous variants \cite{sigal_milder_2022}. This result is also unsurprising in that people are getting weary and tired after two years of avoiding COVID-19.

        \paragraph{Trigger Summary Abstractiveness.}
        The average length of trigger summaries is: $130.9$ characters / $26.9$ words / $1.2$ sentences.
            We measure the abstractiveness of the annotated abstractive summaries of triggers by computing the ROUGE score between the abstractive summaries and the post. we use ROUGE-n precision scores to calculate how abstractive the annotated abstractive summaries are compared to the post. Results are: ROUGE-1: 0.576, ROUGE-2: 0.149, ROUGE-L: 0.392. The results indicate that the trigger summaries are fairly abstractive with respect to the original posts in \dataset{}.

        \paragraph{Topic Variation.}
            To better understand the triggers of each emotion, we use Latent Dirichlet Allocation (LDA) \cite{10.5555/944919.944937} to extract the topics in the trigger summaries of each emotion. The triggers are lower-cased, and punctuation as well as stopwords are removed. We showcase the unigrams corresponding to the most prominent topics in Table \ref{tab:lda}. We observe a clear difference among the topics of triggers behind the emotions. For example, we notice words such as \textit{don}, \textit{vaccinated}, and \textit{mask} in emotions like \textit{anger} or \textit{disgust}, suggesting that the posters are annoyed that people are not masking or vaccinated to prevent the spread of the pandemic. On the other hand, we see words such as \textit{vaccine}, \textit{believe}, and \textit{credible} in \textit{trust}, denoting that the posters believe in the capability of the vaccines to protect them from the virus.

\section{Methods}
We discuss our methods across three main dimensions: emotion detection, summarization, and joint emotion detection and trigger summarization.

\subsection{Separate Models}

\paragraph{Emotion Detection.} To perform emotion detection, we experiment with \textbf{1)} EmoLex, a weak baseline based on the EmoLex lexicon \cite{mohammad2013crowdsourcing}, where words are associated with Plutchik basic emotions. For each post, we assign an emotion $e$ if there exists a word from EmoLex associated with $e$. \textbf{2)} GoEmotions \cite{demszky-etal-2020-goemotions}, which involves training a BERT-large model \cite{devlin-etal-2019-bert} on the GoEmotions dataset, which is composed of sentence-level examples from a general Reddit domain. \textbf{3)} HurricaneEMO \cite{desai-etal-2020-detecting}, the same approach with the model trained on a Twitter disaster dataset. Finally, we use a \textbf{4)} BERT-large model fine-tuned on \dataset{} using the [CLS] token and an additional linear layer to classify the entire post.

\paragraph{Abstractive Summarization.}
We perform abstractive trigger summarization using two backbone models: \textbf{1)} Pegasus \cite{pmlr-v119-zhang20ae} pretrained on Reddit TIFU \cite{Kim:2019:NAACL-HLT} and \textbf{2)} BART \cite{lewis-etal-2020-bart} pretrained on CNN/DailyMail \cite{nips15_hermann}. For each model, we evaluate the summaries with and without fine-tuning on \dataset{}. We employ a separate summarization model for each emotion $e$, which we fine-tune using the abstractive summaries of triggers for $e$. We also experiment with two standard heuristic baselines: i.e., considering the first sentence in the post (\textsc{1-sent}) or the first three sentences (\textsc{3-sent}) as the trigger summary.

\begin{figure}[t]
\centering
\includegraphics[width=0.45\textwidth]{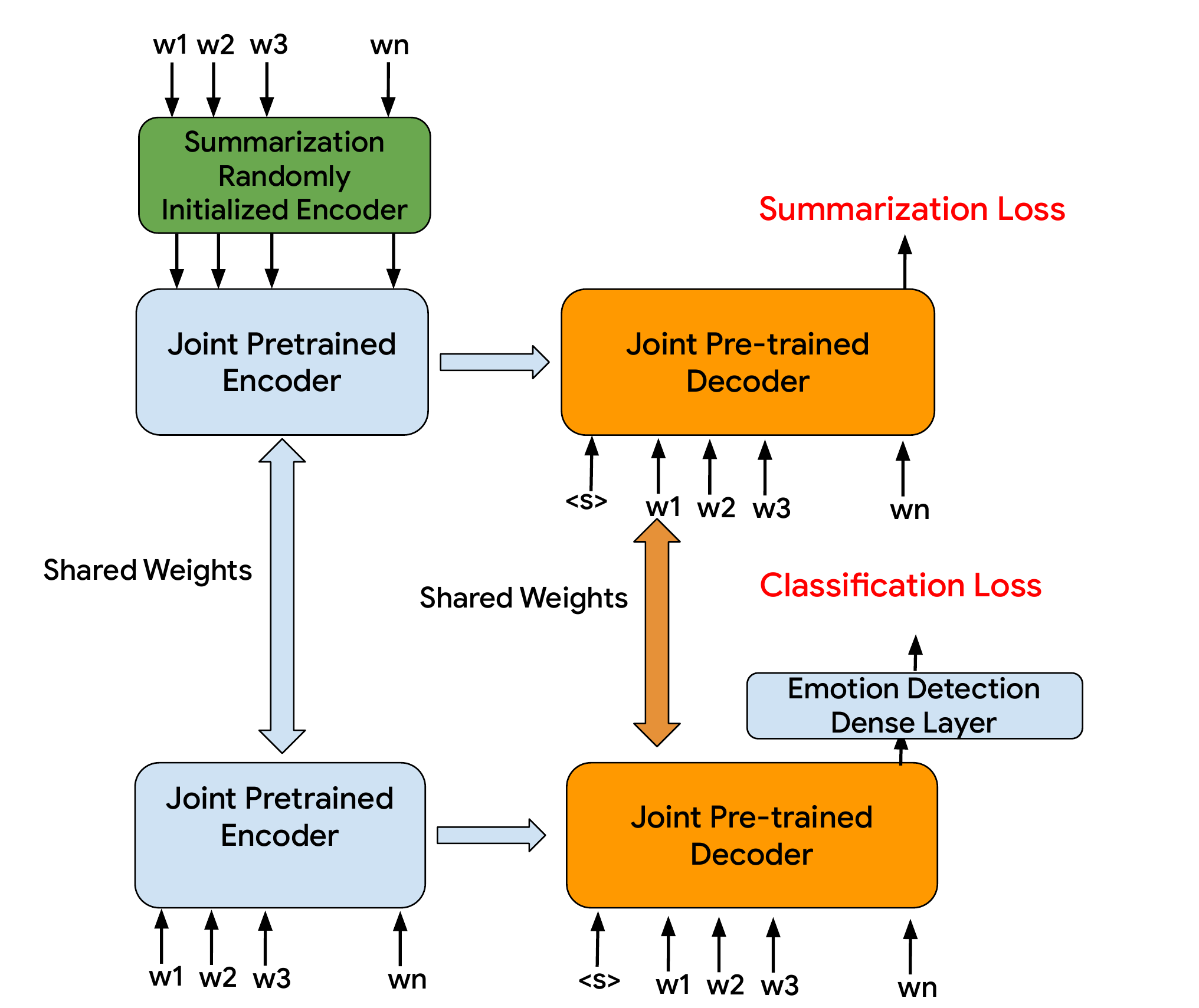}
\caption{Architecture of our joint emotion detection and trigger summarization model.}
\label{fig:joint_architecture}
\end{figure}

\subsection{Joint Emotion Detection and Trigger Summarization}

\looseness=-1
We propose a joint model based on BART that can be trained to simultaneously perform emotion detection and abstractive trigger summarization for a particular emotion $e$ using a  multitasking framework. The model follows the architecture of BART \cite{lewis-etal-2020-bart}, where we add a single linear layer for emotion classification. We show the architecture of our model in Figure \ref{fig:joint_architecture} and detail our training procedure as follows: Given an emotion $e$ and a batch size of $B$, we first sample a positive set of $\frac{B}{2}$ examples: $X_{p}=\{(x_{1}, y_{1}, s_{1}), (x_{2}, y_{2}, s_{2}), ...(x_{\frac{B}{2}}, y_{\frac{B}{2}}, s_{\frac{B}{2}})\}$, where
\begin{equation}
   y_{i} = e \quad | \quad i=1...\frac{B}{2}
\end{equation}

\noindent
and $s_{i}$ is an abstractive summary of the trigger for emotion $y_{i}$ and post $x_{i}$.
Next, we sample a set of negative examples for classification of the same size as follows: $X_{n} = \{(x_{\frac{B}{2} + 1}, y_{\frac{B}{2} + 1}), (x_{\frac{B}{2} + 2}, y_{\frac{B}{2} + 2}), ...(x_{B}, y_{B})\}$, where:
\begin{equation}
   y_{i} \neq e \quad | \quad i=\frac{B}{2}...B
\end{equation}

\noindent
Finally, we use a weighted combinations of the summarization and classification losses to train our model:
\begin{equation}
    L = \lambda * \sum_{i=0}^{B}L_{e}(x_{i}, y_{i}) + (1 - \lambda) * \sum_{i=0}^{\frac{B}{2}}L_{s}(x_{i}, s_{i})
\end{equation}

\noindent
where $L_{e}$ and $L_{s}$ are the regular classification and summarization losses.

\section{Experiments and Results}

\subsection{Experimental Setup} 
We carry out all our experiments on an Nvidia A$5000$ GPU. We use the HuggingFace Transformers \cite{wolf-etal-2020-transformers} library for our model implementations and we will make the code for our methods and data available. We report the performance for emotion detection in terms of F1 and use automatic approaches such as ROUGE \cite{lin-2004-rouge} and BERTScore \cite{DBLP:journals/corr/abs-1904-09675} to evaluate the summarization performance. To enable a fair comparison with the joint model, for summarization, we only consider test examples where the joint model emotion predictions are correct to compute summarization metrics. We run our approaches five times with different model initializations and report average values. We provide extensive details about our hyperparameters, such as batch size or loss weighting $\lambda$ in Appendix \secref{appendix:hyperparameters}. Additionally, we carry out an extensive human evaluation of trigger summaries generated by our BART-FT-JOINT model and a general BART \cite{lewis-etal-2020-bart} model trained on CNN/DailyMail.

\begin{table}[!t]
\setlength{\tabcolsep}{3pt}
\centering
\tiny
\begin{tabular}{r|cccccccc}
 & \textsc{agr} & \textsc{dsg} & \textsc{fer} & \textsc{joy} & \textsc{sdn} & \textsc{trs} & \textsc{anc} & \textsc{avg} \\
\toprule
\textsc{emolex} & $35.6$ & $20.5$ & $56.7$ & $48.7$ & $42.5$ & $13.5$ & $17.8$ &  $33.6$\\ 
\textsc{goemotions} & $45.4$ & $20.1$ & $65.3$  & $50.4$ & $58.3$ & $15.1$ & $41.3$ & $42.2$\\
\textsc{hurricaneemo} & $37.1$  & $16.8$ & $58.3$ & $45.2$ & $60.7$ & $17.2$ &  $43.7$ & $39.9$\\
\textsc{bert-large} & $68.1$ & $20.2$ & $86.8$ & $54.2$ & $69.5$ & $20.3$ & $64.5$ & $54.8$\\
\midrule
\textsc{bart-ft-joint} & $69.5^{\dagger}$ & $20.6$ & $87.8^{\dagger}$ & $54.7$ & $71.3^{\dagger}$ & $20.8$ & $65.9^{\dagger}$ & $55.8^{\dagger}$ \\
\bottomrule
\end{tabular}
\caption{Results of our models in terms of F1 on emotion detection. We report the average performance of five independent runs. We use bootstrap statistical significance$^{\dagger}$ testing over \textsc{bert-large} with $p < 0.05$ and $200$ samples of size $10\%$ of the test set.}
\label{tab:emotion_detection}
\end{table}

\begin{table*}[!htbp]
\setlength{\tabcolsep}{6pt}
\centering
\small
\resizebox{16cm}{!}{%
\begin{tabular}{r|cc|cc|cc|cc|cc|cc|cc}
 & \multicolumn{2}{c}{\textsc{anger}} & \multicolumn{2}{c}{\textsc{disgust}} & \multicolumn{2}{c}{\textsc{fear}} & \multicolumn{2}{c}{\textsc{joy}} & \multicolumn{2}{c}{\textsc{sadness}} & \multicolumn{2}{c}{\textsc{trust}} & \multicolumn{2}{c}{\textsc{anticipation}} \\
& R-L & BERTSc & R-L & BERTSc & R-L & BERTSc & R-L & BERTSc & R-L & BERTSc & R-L & BERTSc & R-L & BERTSc \\
 \toprule
\textsc{1-sent} & $0.121$ & $0.575$ & $0.112$ & $0.545$ & $0.122$ & $0.528$ & $0.103$ & $0.518$ & $0.115$ & $0.506$ & $0.118$ & $0.537$ & $0.119$ & $0.507$ \\ 
\textsc{3-sent} & $0.142$ & $0.598$ & $0.129$ & $0.562$ & $0.153$ & $0.535$ & $0.154$ & $0.537$ & $0.134$ & $0.517$ & $0.152$ & $0.548$ & $0.142$ & $0.527$ \\ 
\textsc{pegasus} & $0.164$ & $0.594$ & $0.141$ & $0.560$ & $0.161$ & $0.548$ & $0.155$ & $0.536$ & $0.153$ & $0.562$ & $0.151$ & $0.546$ & $0.153$ & $0.542$ \\ 
\textsc{bart} & $0.161$ & $0.587$ & $0.138$ & $0.558$ & $0.164$ & $0.529$ & $0.149$ & $0.551$ & $0.157$ & $0.559$ & $0.158$ & $0.571$ & $0.164$ & $0.558$ \\ 
\textsc{pegasus-ft} & $0.185$ & $0.681$ & $0.155$ & $0.713$ & $0.199$ & $0.739$ & $0.158$ & $0.683$ & $0.173$ & $0.705$ & $0.164$ & $0.663$ & $0.193$ & $0.736$ \\
\textsc{bart-ft} & $0.190$ & $0.705$ & $0.159$ & $0.695$ & $0.206$ & $0.748$ & $0.165$ & $0.699$ & $0.177$ & $0.718$ & $0.162$ & $0.653$ & $0.198$ & $0.749$ \\
\midrule
\textsc{bart-ft-joint} & $0.190$ & $0.701$ & $0.158$ & $0.706$ & $0.203$ & $0.729$ & $0.163$ & $0.694$ & $0.175$ & $0.713$ & $0.165$ & $0.659$ & $0.196$ & $0.746$ \\
\bottomrule
\end{tabular}
}
\caption{Results of our models in terms of ROUGE-L and BERTScore on the trigger summarization subtask of emotion detection and trigger summarization. We report the average performance of five independent runs.
}
\label{tab:summarization}
\end{table*}

\subsection{Results}

\paragraph{Emotion Detection.}
We show the F1s obtained using our models on emotion detection in Table \ref{tab:emotion_detection}. First, we observe that our lexicon-based EmoLex approach performs poorly compared to other methods. We also note that approaches trained outside our domain lag behind considerably compared to approaches trained on our data. Specifically, a BERT large model trained on our data outperforms the GoEmotions model by as much as $23\%$ in F1 on anger and $28\%$ in fear. We observe the same trend for models trained on hurricane disasters, which decrease the performance by $38\%$ on fear and $9\%$ on joy. This result indicates that models trained on natural disasters generalize poorly to Covid-19, further emphasizing the uniqueness of our dataset. We also note that our BART-FT-JOINT model, which is trained on our data to perform both detection and summarization obtains an average improvement of $1\%$ over the BERT-large model.

\paragraph{Trigger Summarization.}
We show in Table \ref{tab:summarization} the results obtained in terms of ROUGE-L and BERTScore on the summarization task. First, we note that basic approaches such as 1-SENT or 3-SENT, which select the first sentences in a post as the trigger summaries, perform similarly to general summarization models like the Pegasus model trained on Reddit TIFU or the BART trained on CNN/DailyMail. This result highlights the distinct nature of our trigger summarization task, which bears very few similarities with a general summarization task. Fine-tuning these models on our data, however, brings substantial improvements. We see improvements as large as $18\%$ in terms of BERTScore by fine-tuning a BART model on anger and $19\%$ on anticipation. Our fine-tuned models also consistently outperform the baselines in ROUGE-L. For instance, our fine-tuned Pegasus obtains an improvement of $4.2\%$ ROUGE-L on fear and $2\%$ on sadness. We note that applying our joint model results in no loss of performance across all emotions.

We emphasize that in practice, generating trigger summaries and detecting emotions using a joint model has various advantages over single-task approaches, such as reduced memory footprint (i.e., by using a single model) and reduced inference time. Moreover, our approach improves the performance in emotion detection.

\begin{table}[!t]
\setlength{\tabcolsep}{4pt}
\centering
\small
\adjustbox{max width=\columnwidth}{%
\begin{tabular}{r|ccccc}
\\
\textsc{metric} & \emph{Coherence} & \emph{Consistency} & \emph{Fluency} & \emph{Relevance} & \emph{Extractiveness} \\
\toprule
\textsc{bart} & $4.947$ & $5.000$ & $4.974$ & $2.158$ & $4.970$\\
\textsc{bart-ft-joint} & $4.262$ & $3.548$ & $4.286$ & $4.048$ & $2.530$\\
\bottomrule
\end{tabular}}
\caption{Results of our trigger summary human evaluation procedure along four quality assessment dimensions.}
\label{tab:evaluation_results}
\end{table}

\subsection{Human Evaluation of Model Summaries} 
\begin{figure}[t]
    \centering
    \includegraphics[width=\linewidth]{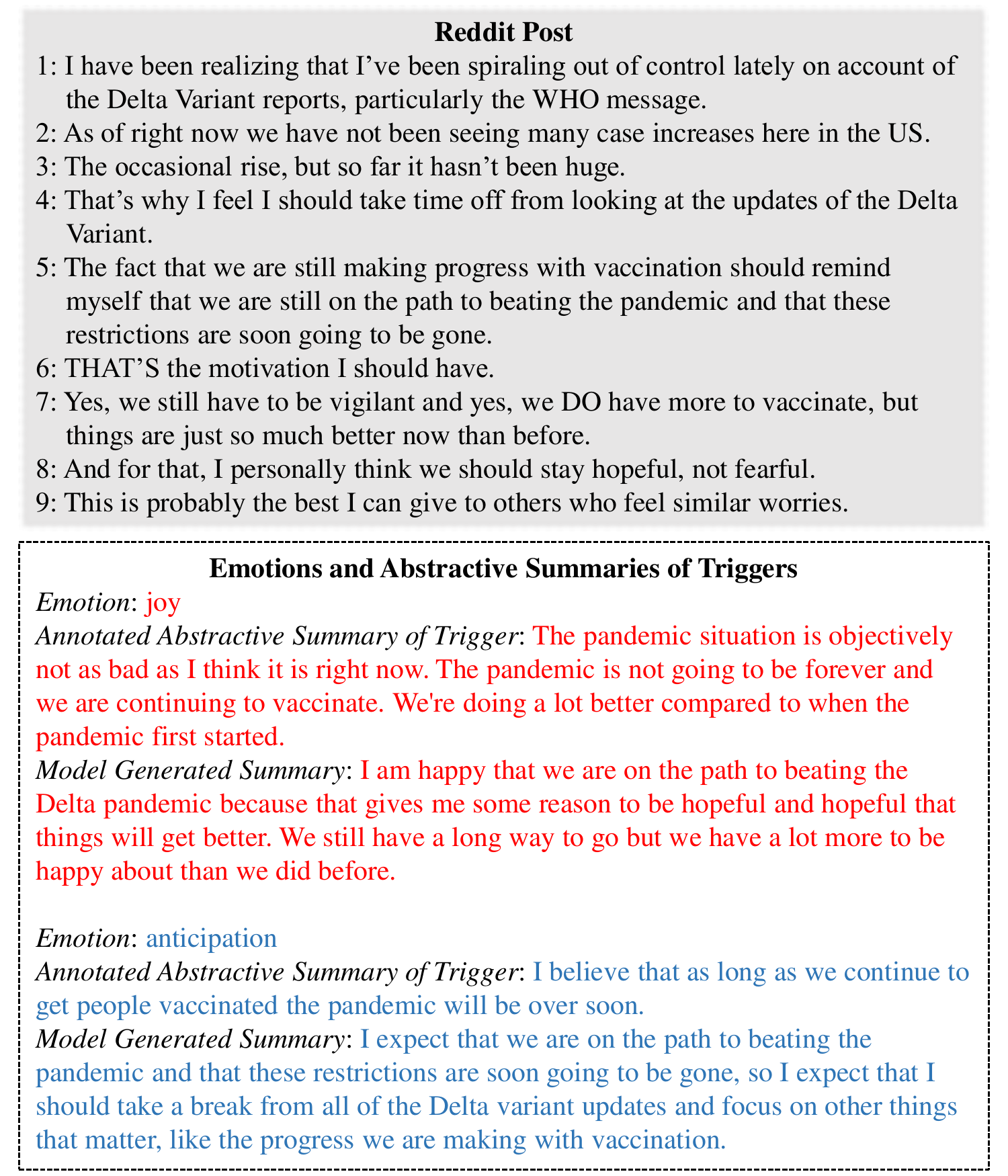}
    \caption{Human Evaluation Example.}
    \label{fig:human_eval_example}
\end{figure}

We perform human evaluation and qualitative analysis of our model-generated trigger summaries to measure the overall quality and compare our BART-FT-JOINT model against a general BART summarization model.

Following \citet{fabbri-etal-2021-summeval}, we instruct two expert annotators with linguistics expertise to grade with a score from $1$ to $5$ (where $1$ is the lowest score) $21$ trigger summaries generated by our joint model (three per emotion) along four dimensions: \emph{Coherence}, \emph{Consistency}, \emph{Fluency}, and \emph{Relevance}. Coherence refers to the collective quality of all the sentences in the summary and consistency measures the factual alignment between the summary and the summarized source. Next, we evaluate the quality of individual sentences from the post using fluency and measure how well the summary captures the emotion triggers through relevance. To offer a better understanding of these metrics, we detail them further in Appendix \secref{appendix:human_eval}. Additionally, we also evaluate the summaries for the amount of \emph{Extractiveness} (i.e., the amount of information copied from the original post).

We show the evaluation results in Table \ref{tab:evaluation_results}. The reported metrics are the average scores of the two individual annotators' scores. We measure the agreement between the two annotators by computing the average score differences between their responses.

Evaluation of BART-FT-JOINT yields a small average difference of $0.690$, indicating that the two annotators have good agreement on the assigned scores. The generated summaries have a good quality, with an average score of $4$. We also note that the lowest score of $3.548$ is obtained on consistency, indicating that the model can introduce non-factual details, and emphasize that our summarization model performs well identifying triggers, where it obtains a score of $4.048$. To offer additional insights into the summaries generated by our joint model, we show an example in Figure \ref{fig:human_eval_example}. The post is annotated as joy and anticipation, and we provide both the gold and the model generated summaries. The summary for joy emotion is extremely effective capturing the trigger; i.e., the progress towards beating the Delta variant. However, we also note some model errors, such as the repetition of the word ``hopeful''. The annotators indicate that the model outputs tends to be two sentences long and the overall quality is good. Besides scoring the summaries, we also instruct annotators to spot such mistakes of the model in order to identify potential areas of improvement. We detail our findings in Appendix \secref{human_eval_findings}.

As mentioned, we also provide in Table \ref{tab:evaluation_results} the Likert scoring of the generic summarization model by linguistic experts. Inspection of the data reveals that the generic summaries tend to be word-to-word extractive of the original post, leading to high scores in coherence, consistency, and fluency. However, the generic summaries perform badly in terms of relevance, suggesting that the models are not capturing the triggers of the emotions. This is also reflected in the low BERTScore performance for the generic models.

\section{Conclusion}
We propose a new task entitled \textit{emotion detection and trigger summarization}, which aims to jointly detect perceived emotions in text and summarize the events as well as their appraisals that trigger each emotion. To address the task, we introduce \dataset{}, a dataset of $1,883$ English Reddit posts on COVID-19 annotated with emotions and abstractive summaries of their triggers. Experiments using our proposed joint model on the dataset reveal that \dataset{} is vital resource for training models to capture emotions and their triggers in text. Our thorough evaluation of model-generated summaries emphasize that \dataset{} is a challenging benchmark, and our error analysis indicates potential areas of improvements (e.g., improving the factuality of the summaries).

\section*{Limitations}
This work presents a new dataset to address the task of detecting perceived emotions and summarizing their triggers in text. While picking the COVID-19 as our topic enables meaningful, real-world applications and allows us to access emotionally rich text, the emotion labels in \dataset{} are highly unbalanced: negative emotions such as \textit{fear} and \textit{anger} are more prevalent. This makes it particularly challenging to train emotion detection and summarization models on emotions with few examples (e.g., \textit{trust} and \textit{joy}). Moreover, due to the lack of controllability and interpretability of end-to-end summarization models, we acknowledge the potential risks of generating biased or inappropriate trigger summaries for certain posts. In particular, our results revealed consistency and factuality issues that exist in modern abstractive summarization systems.

\section*{Acknowledgements}
We thank Desmond Ong and the anonymous reviewers for their valuable comments and feedback, which helped improve our paper. We also thank our annotators for their dedication and hard work. We thank John Henry Cruz for his help with Reddit data.
This research is partially supported by Good Systems,\footnote{\url{https://goodsystems.utexas.edu/}} a UT Austin Grand Challenge to develop responsible AI technologies, and NSF grants IIS-2145479, IIS-2107524, IIS-2107487, and BigData-1912887. We acknowledge the Texas Advanced Computing Center (TACC)\footnote{\url{https://www.tacc.utexas.edu}} at UT Austin and AWS for many of the results within this paper.

\bibliography{custom}
\bibliographystyle{acl_natbib}

\appendix
\section{Dataset Examples}\label{appendix:dataset-examples}
\begin{figure*}[!htbp]
    \centering
    \includegraphics[width=\textwidth]{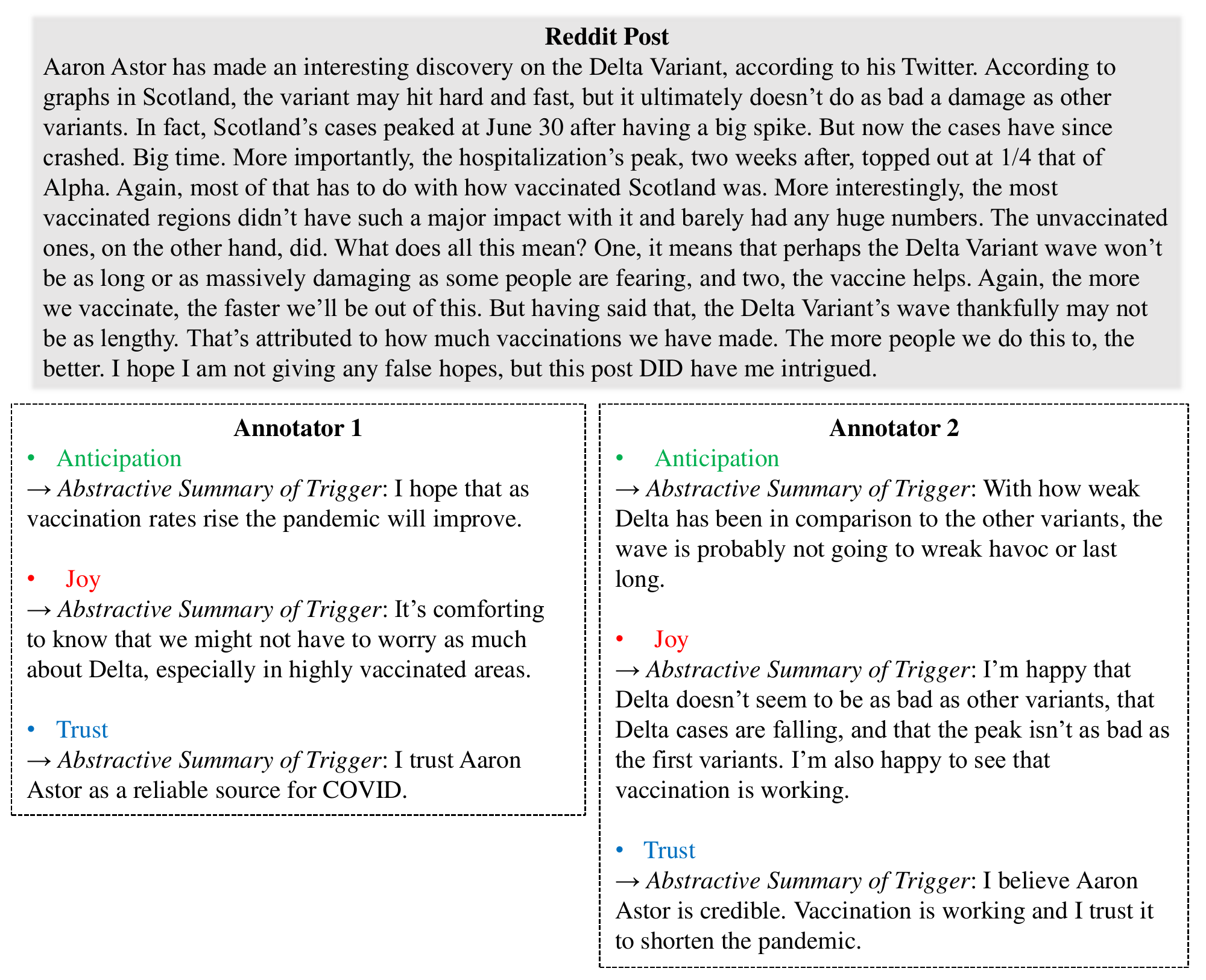}
    \caption{Example of \dataset{}.}
    \label{fig:dataset_example}
\end{figure*}

    An example of \dataset{} is shown in Figure \ref{fig:dataset_example}. This example includes annotations from both annotators. Annotations for different emotions are in distinct colors.

\section{Data Curation Details}\label{appendix:data-curation}
    Here we detail the preprocessing procedure over the source data. We preprocess the source data using regular expressions. As the first step, we tokenize posts into individual words. Specifically, we apply the following regular expressions in combination with the NLTK word\_tokenize package to tokenize posts into words:
        \begin{lstlisting}
re.sub("\s+"," ", post)
re.sub(r'(?<=[.,!?:])(?=[^\s])', r' ', post)
re.sub(r'\s([?.!,:"](?:\s|$))', r'\1', post)
nltk.tokenize.word_tokenize(post)
        \end{lstlisting}
    
    Then we exclude punctuation from the tokenized posts and filter the posts that are 50-400 tokens long. Finally, we mask web links by substituting them into \texttt{[url]} tokens using the following regular expressions:
    
        \begin{lstlisting}
pandas.Series.str.replace(r'http\S+', '[url]').str.strip()
pandas.Series.str.replace(r'''(?i)\b((?:https?://|www\d{0,3}[.]|[a-z0-9.\-]+[.][a-z]{2,4}/)(?:[^\s()<>]+|\(([^\s()<>]+|(\([^\s()<>]+\)))*\))+(?:\(([^\s()<>]+|(\([^\s()<>]+\)))*\)|[^\s`!()\[\]{};:'".,<>?«»“”‘’]))''', '[url]').str.strip()
        \end{lstlisting}

\section{Annotation Instructions}\label{appendix-subsec:annotation-instructions}

\begin{figure*}[th]
    \centering
    \adjustbox{max height=\textheight}{\includegraphics{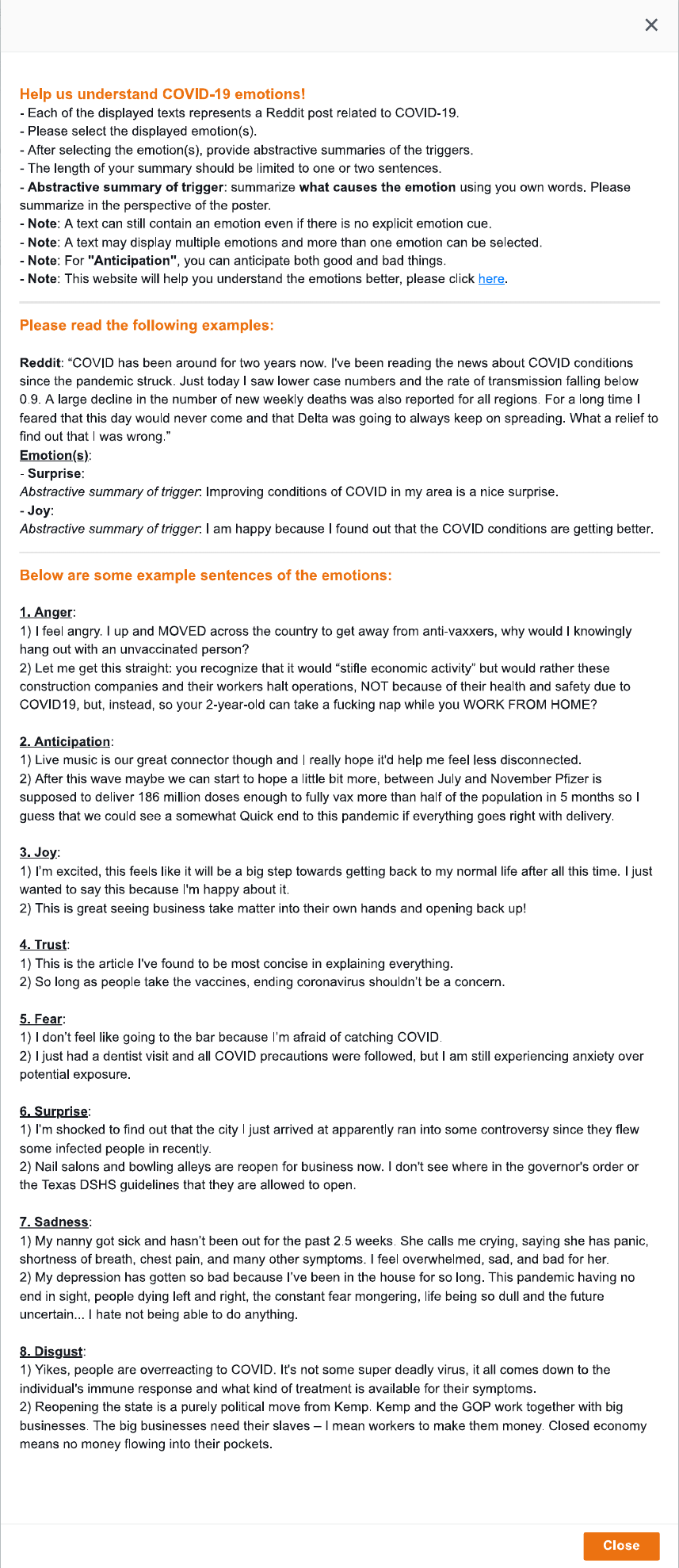}}
    \caption{Annotation instructions (always shown before annotating).}
    \label{fig:annotation-task-instructions}
\end{figure*}

        Comprehensive instructions are provided to the annotators, as demonstrated in Figure \ref{fig:annotation-task-instructions}. Note that the instruction page pops up as a modal before every annotation, so as to remind the annotators of the task framework. We also ask the annotators to pay special attention to a few principles as follows. For the emotion annotations, we ask annotators to follow the emotion guidelines on the Six Seconds website\footnote{\url{https://www.6seconds.org/2020/08/11/plutchik-wheel-emotions/}} and interpret \textit{anticipation} as (good or bad) \textit{expectancy} \citep{plutchik-1958}. For the trigger annotations, we instruct annotators to annotate summaries containing \textit{triggers} that lead to the emotion instead of sentences expressing the \textit{emotion} itself.
        
\begin{figure*}[htp]
    \centering
    \includegraphics[width=\textwidth]{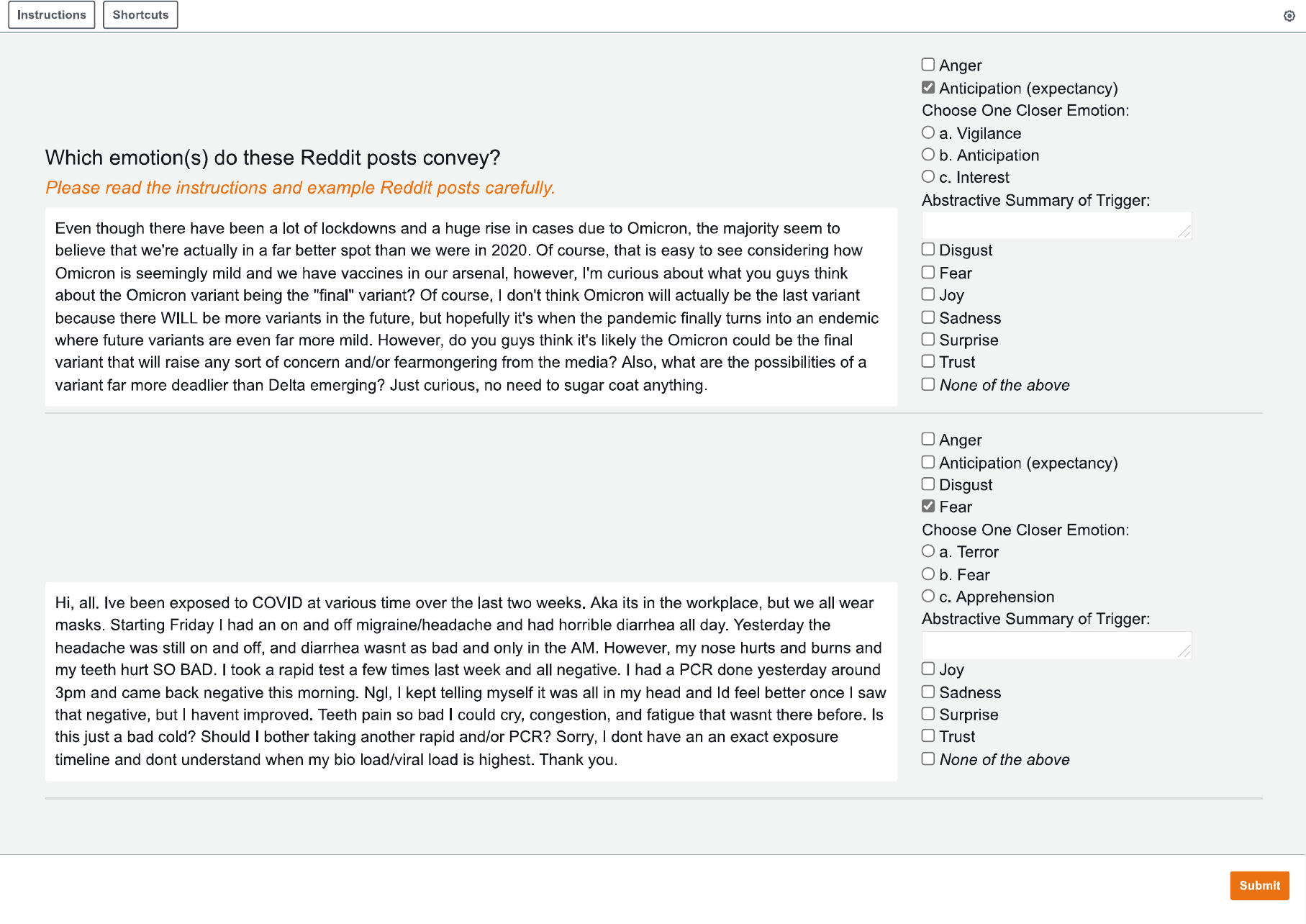}
    \caption{The annotation task layout of an example hit on the Amazon Mechanical Turk.}
    \label{fig:annotation-task-layout}
\end{figure*}

        The layout of our annotation task is shown in Figure \ref{fig:annotation-task-layout}.

\section{Hyperparameters}
\label{appendix:hyperparameters}
In this section, we detail the hyperparameter search space and the final hyperparameters used by our joint BART-FT-JOINT model, which were chosen based on the best validation performance. Specifically, we show the values for the learning rate, batch size and multitasking loss weighting term $\lambda$ in Table \ref{tab:hyperparameters_used}. In terms of search space, we tried batches in the range $4 \rightarrow 64$ and learning rates in the range $1e-5 \rightarrow 9e-5$ with a step of $1e-5$. We also search a suitable $\lambda$ in the range $0.1 \rightarrow 0.9$. We decode our summaries using beam search decoding and a beam size of $4$. Training BART-FT-JOINT model on our A$5000$ GPU takes \textasciitilde$1$ hour to complete for each emotion.

\section{Human Evaluation Instructions}\label{appendix:human_eval}
\begin{figure*}[ht]
    \centering
    \includegraphics[height=0.8\textheight]{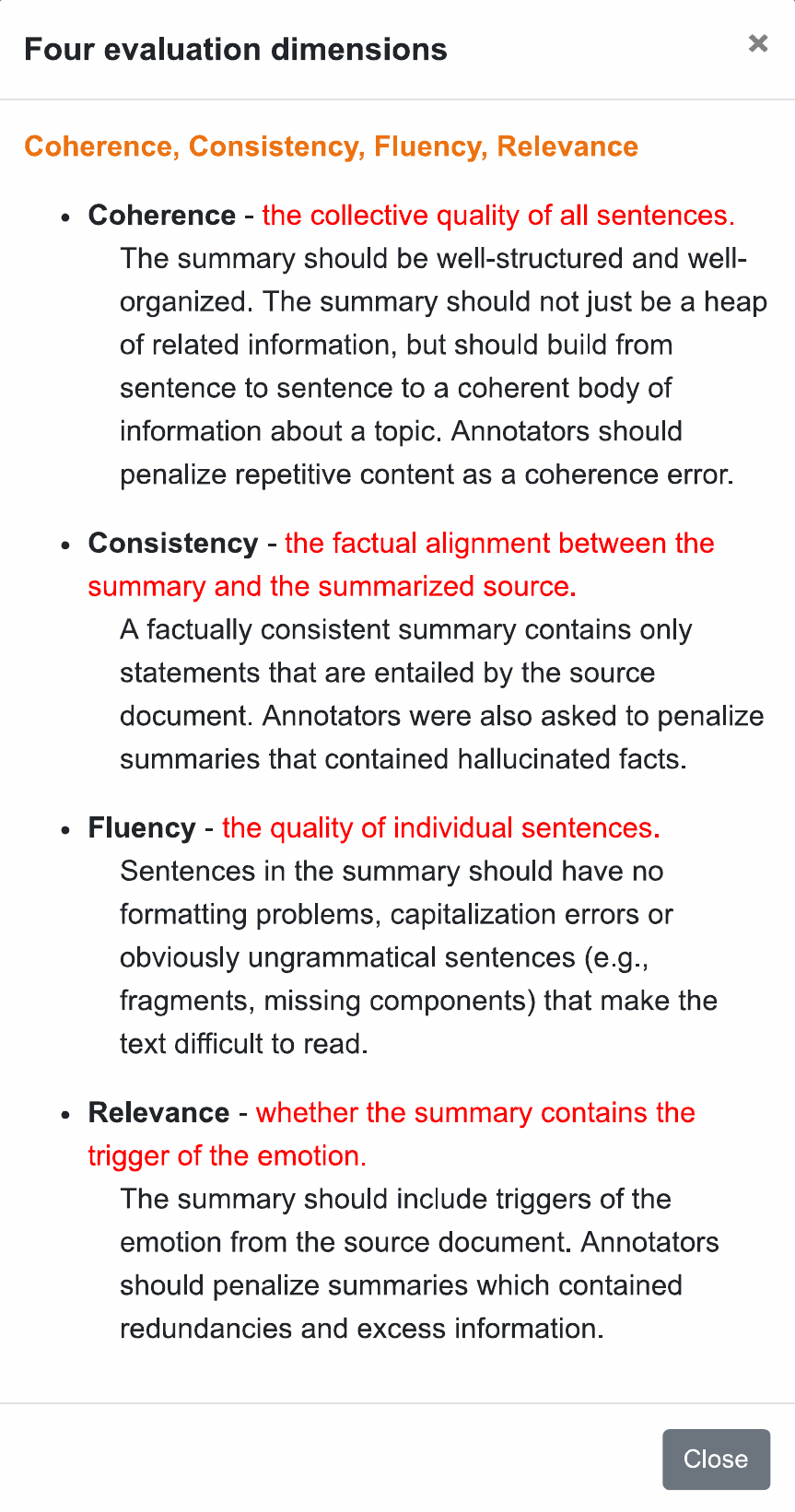}
    \caption{Human Evaluation Instructions.}
    \label{fig:human_eval_instructions}
\end{figure*}

We provide the detailed instructions for human evaluation in Figure \ref{fig:human_eval_instructions}.

\section{Human Evaluation Summary Errors}
\label{human_eval_findings}

We instructed our expert human evaluators to find potential areas of improvement of our BART-FT-JOINT summarization model by identifying frequent errors the model makes. In this section, we analyze our findings and present a few examples in Table \ref{tab:human_identified_common_errors}. Specifically, the annotators pointed out four main model errors: \textbf{1)} Non-factual relative clauses; \textbf{2)}  Model summary includes information in the summary that is not discussed in the post; \textbf{3)} At least a few sentences in the model summary are formatted to make the text difficult to read; and \textbf{4)} The overall model summary is not well-structured.

\begin{table*}[!htbp]
\setlength{\tabcolsep}{6pt}
\centering
\small
\resizebox{16cm}{!}{%
\begin{tabular}{r|cccccccc}
 & \textsc{anger} & \textsc{disgust} & \textsc{fear} & \textsc{joy} & \textsc{sadness} & \textsc{trust} & \textsc{anticipation} \\
\toprule
Batch Size & $32$ & $32$ & $32$ & $8$ & $32$ & $8$ &  $16$ \\
Learning Rate & $2e-5$ & $4e-5$ & $5e-5$ & $3e-5$ & $3e-5$  & $5e-5$ & $3e-5$ \\
Loss weight $\lambda$ & $0.1$ & $0.1$ & $0.2$ & $0.1$ & $0.1$ & $0.2$ & $0.1$\\
\bottomrule
\end{tabular}
}
\caption{Hyperparameters of our BART-FT-JOINT model.
}
\label{tab:hyperparameters_used}
\end{table*}

\begin{table*}[t]
\vspace{-1cm}
\tiny
\centering
\begin{tabular}{p{8.8cm}|p{4.4cm}}
\toprule
\textsc{Post} & \textsc{Summary} \\
\midrule
\multicolumn{2}{c}{\textbf{Non-factual relative clauses}} \\
\midrule
I am visiting family this week. All of my family members who are eligible for the vaccine, including myself. have gotten vaccinated. The only people who arent are my niece and nephew, because they are only 4 and 1 years old. I wanted to see an old friend from high school, but I found out that this friend did not get vaccinated. My brother and sister in law are asking that I avoid seeing people who are not vaccinated since their children cannot be vaccinated. Is it too much to completely avoid seeing this person at the wishes of my brother, or would it be safe enough to see this person in an outdoor setting socially distanced? I know this is a stupid question, but Im curious what you all would do. & I'm looking for any advice about whether it's safe to see a friend who isn't vaccinated \textbf{when their children can't be vaccinated}, or whether I should follow the wishes of my brother and sister-in-law, who want me to avoid this person.
.\\
\midrule
My country is in a third wave. Its never ending. Im sure this post will get removed because Im too depressed/depressing for Reddit. Im broke because of the pandemic. Im struggling to work because my anxiety gives me physical symptoms. My friends dont talk to me anymore, or when I do talk to them they just tell me to see a therapist even though I already do (and can barely afford it). Im thinking about stopping eating so I can afford therapy. Theres no point to any of this. Every death that is about to happen here could have been prevented, and no one cares and they call me crazy. Im tired of the endless hurt. & I expect that this post will be removed from Reddit \textbf{because I'm too depressed to post} it because I expect that no one will want to read it and everyone will think that I'm crazy for thinking that I should stop eating so that I can afford to see a therapist.
.\\
\midrule
\multicolumn{2}{c}{\textbf{Model Summary includes information in the summary not discussed in the post}} \\
\midrule
israel imposed their mask mandate, despite being one of the most vaccinated countries. i feel like this will never end and I don't need stupid replies like ``hang in there'' or ``it will be okay.'' and don't remove this post because it ``causes anxiety.'' I'm not. I'm simply worried that we'll never get back to normal.
 & I'm afraid that we're never going to be able to get back to normal after COVID, because Israel has a mandate and Israel is one of the most vaccinated countries in the world, and \textbf{that's putting a strain on our health systems} and on our mental health.
\\
\midrule
I am fully vaccinated with the Moderna shot, and have been getting back to my regular life. I live in a state (MA) with extremely high vaccination rates, but nonetheless Im concerned about the Delta variant. Ive been hearing stories of breakthroughs of the variant in fully vaccinated people, so I am concerned. At the same time, Im extremely tired of this. I feel selfish for saying this, Ive done all I can do to protect myself and others against COVID, and I want to continue to live my life after over a year of taking precautions. Anyone else feel the same?
 &  I am disgusted to see myself being selfish because I have done all I can to prevent COVID and I want to live my life as if I had not had any COVID side effects at all since I have taken every precaution possible to prevent the COVID variant. \\
\midrule
\multicolumn{2}{c}{\textbf{Some sentences in the model summary are formatted to make the text difficult to read}} \\
\midrule
So the Vaccine team in Iceland is taking a summer holiday for a month that extends over the time when I was suppose to get my second Astra Zeneca shot. They offered me to get it sooner but I heard it will decrease it's effectiveness by allot. Should I get the shot 7 weeks after my first shot or should I wait until they come back and get it at least 15 weeks after after my first shot. Iceland has stopped all restrictions so i am a bit nervous. & I trust that the vaccine will help protect me from catching COVID and \textbf{I'll get it as soon as I can get it}. I trust that COVID \textbf{will do what it's supposed to do and do it it will do its best} to protect me and my family from COVID.
\\
\midrule

Is anyone else experiencing bad post-outing anxiety? Ive been trying to push myself out of my comfort zone (and sometimes I even get excited to) and so I get ahead of myself and leave my house. No bars or clubs, but I did attend an outdoor gathering thats weighing heavy on my mind. While Im out, Ive surprisingly found Im quite bored when I leave my house, but the real problem comes the day afterwards. I sit and think was that too soon? Can I re-enter my bubble now that people have seen me? Am I a hypocrite? And these questions flow through my brain in a never ending sequence. Feeling that Ive been perceived by others and I cant take it back feels unbearable, yet I took the decision to leave my house so I then encounter feelings of embarrassment, guilt and shame. Not to mention the obvious fear of the delta variant, and overall uncertainty over cdc recommendations. (I wish someone could spell out a good plan for reintegration besides - rip off the mask and live!) Everyone I know has returned to life normally and Im here in a weird limbo. I know a lot of people in this struggle with just leaving the house, but is anyone else out there struggling with how they feel once they do? &  I find it hard to leave the house and it surprises me when \textbf{I find out that I'm not feeling the same way when I do}. I feel embarrassed and ashamed that I've been seen out in public and that I can't go back and change what I've done. \\
\midrule
\multicolumn{2}{c}{\textbf{ The overall model summary was not well-structured.
}} \\
\midrule
I've recently heard of stories that people who are getting mildly sick after being vaccinated are still coming down with long covid. People on the covid long hauler subreddit are saying that we're going to have a huge problem on our hands because no vaccine protects against long covid....This isn't going to end, isn't it? Right as I think I've tasted freedom, I find out information that makes me want to stay inside forever like a hermit... & I feel sad and hopeless because I think I've tasted freedom and then I find out more information that makes me want to stay inside like a hermit. I wish I could just be free from this virus for a while but it doesn't look like that will be possible.
\\
\midrule
This makes me really just not want to go out and about again.... I've been on this sub for a while and posted a lot. More or less this pandemic has crushed my mental health and with having some health issues makes me really hesitant to do anything. I was finally getting my life back a little and this Delta variant makes me want to go back to old habits and just stay home and see no one... I really am at a loss of what to do and am feeling super overwhelmed. & I'm at a loss for what to do and don't know what I can do to get back on track with my health issues, so I just want to go back to my old ways and stay home and see no one. I was finally getting my life back before the pandemic hit.
 \\
\bottomrule
\end{tabular}
\caption{Example of common model errors identified by the expert evaluators.}
\label{tab:human_identified_common_errors}
\end{table*}

\end{document}